\documentclass[preprint,12pt]{elsarticle}

\usepackage{amssymb}
\usepackage[utf8]{inputenc}
\usepackage[colorlinks,citecolor=red,urlcolor=blue,bookmarks=false,hypertexnames=true]{hyperref} 
\usepackage{graphicx}
\usepackage{tikz}
\usetikzlibrary{external}
\usepackage{caption}
\usepackage{subcaption}
\usepackage{makecell}
\usepackage{amsmath}
\usepackage{amsmath,amsfonts}
\usepackage{mathabx}
\usepackage{amsthm}%
\usepackage{mathrsfs}%
\usepackage{pifont}
\usepackage{textcomp}

\begin{document}

\begin{frontmatter}




\title{Stacked ensemble-based mutagenicity prediction model using multiple modalities with graph attention network}

\author[inst1]{Tanya Liyaqat}

\affiliation[inst1]{organization={Department of Computer Engineering},
            addressline={Jamia Millia Islamia},
            country={New Delhi, India}}

\author[inst1]{Tanvir Ahmad}

\author[inst1]{Mohammad Kashif}
\author[inst2]{Chandni Saxena}

\affiliation[inst2]{organization={The Chinese University of Hong Kong},
            state={Hong Kong SAR},
            country={China}}

\begin{abstract}
Mutagenicity is a concern due to its association with genetic mutations which can result in a variety of negative consequences, including the development of cancer. Earlier identification of mutagenic compounds in the drug development process is therefore crucial for preventing the progression of unsafe candidates and reducing development costs. While computational techniques, especially machine learning models have become increasingly prevalent for this endpoint, they rely on a single modality. In this work, we introduce a novel stacked ensemble based mutagenicity prediction model which incorporate  multiple modalities such as simplified molecular input line entry system (SMILES) and molecular graph. These modalities capture diverse information about molecules such as substructural, physicochemical, geometrical and topological. To derive substructural, geometrical and physicochemical information, we use SMILES, while topological information is extracted through a graph attention network (GAT) via molecular graph. Our model uses a stacked ensemble of machine learning classifiers to make predictions using these multiple features. We employ the explainable artificial intelligence (XAI) technique SHAP (Shapley Additive Explanations) to determine the significance of each classifier and the most relevant features in the prediction. We demonstrate that our method surpasses state-of-the-art methods on two standard datasets across various metrics. Notably, we achieve an area under the curve of 95.21\% on the Hansen benchmark dataset, affirming the efficacy of our method in predicting mutagenicity. We believe that this research will captivate the interest of both clinicians and computational biologists engaged in translational research. The code used in this research is available through the provided link~\url{https://github.com/Believer604/STEM.git}

\end{abstract}



\begin{keyword}
Mutagenicity\sep Drug discovery\sep Stacked ensemble \sep Graph Attention Network \sep Multiple modalities
\end{keyword}

\end{frontmatter}


\section{Introduction}
\label{sec:Intro}
Anticipating compound toxicity in the early phases of drug discovery is essential for expediting the process and mitigating safety-related setbacks during drug development\cite{segall2014addressing}. Within the spectrum of toxicological effects, 
particular emphasis is placed on the genotoxic potential of compounds. Genotoxicity is the negative impact of a chemical substance that damages a  genetic material of cells, especially its DNA\cite{phillips2009genotoxicity}. Genotoxic chemicals cause mutations or other structural changes in genetic material, which result in a number of detrimental outcomes, including the development of cancer. Mutagenicity is a type of genotoxicity that causes mutations in the DNA sequence which may result in heritable modifications to the genetic code. Thus, mutagenic testing is essential for determining the safety of compounds, particularly those that may be ingested, inhaled or come into reach of living beings\cite{brambilla2009genotoxicity}. 
The Ames test\cite{ames1973improved} is a popular bacterial assay employed to assess the potential of a chemical to induce mutagenesis. It is named after its creator, Dr. Bruce Ames. 
\emph{Salmonella typhimurium} strains often shortened to \emph{S. typhimurium} — are the main source of bacteria used in the Ames test. These bacterial strains are auxotrophic mutants, meaning they are unable to synthesis a specific necessary amino acid. Because of this, the bacteria are more vulnerable to mutagenesis. If the chemical causes mutations, it might reverse the auxotrophic mutation in the bacteria, enabling them to grow on a medium that doesn't contain the amino acid they were previously unable to manufacture. This reversal is a sign of mutagenicity\cite{mortelmans2000ames}. \\
\indent However, since this approach fails to keep up with the exponential growth in the number of compounds, researchers and regulatory bodies have developed screening techniques that can effectively detect substances that may be harmful without the need for costly toxicity testing\cite{muster2008computational}. In this quest, quantitative structure–activity relationship (QSAR) approaches for predicting Ames mutagenicity have proven to be effective\cite{honma2019improvement,honma2020assessment}. \\
\indent QSAR methods  entail the development of models that  establish correlations between the structural properties of chemical substances with their mutagenic activity allowing for the anticipation of mutagenicity without undergoing lengthy experimental testing. QSAR models are one of the in silico methodologies that are recognized as useful tools for the assessment of mutagenic contaminants by the International Conference on Harmonization (ICH) M7\cite{chandramore2023regulatory}. The guideline suggests that in silico methods can be used in conjunction with experimental studies to assess the mutagenic potential of substances. In recent years, the technique of mutagenicity prediction based on molecular structures has steadily gained attention. For mutagenicity prediction, there are numerous rule-based and statistical-based expert systems. The most notable of these are ProTox-II\cite{banerjee2018protox}, CASEUltra\cite{saiakhov2013effectiveness}, and DEREK\cite{marchant2008silico}. Rule-based approaches can provide transparent and interpretable models. But they may struggle with capturing complex relationships and may not generalize well to new, unseen data as they are often dependent on the quality and relevance of the rules derived from expert knowledge\cite{patlewicz2008evaluation}. On the contrary, statistical-based methods can capture complex, nonlinear relationships and are capable of generalizing to new data. However, they may lack interpretability compared to rule-based approaches. The “black-box” nature of some complex models can make it challenging to comprehend the underlying reasons for predictions\cite{marchant2008silico}.\\
\indent Nowadays, machine-learning (ML) methods are frequently employed. 
Dataset with details on chemical compound structures and associated toxicological effects are used to train machine learning models\cite{baskin2018machine}. Molecular descriptors, molecular fingerprints or high-order latent representative features are derived from the chemical structures of compounds to capture information regarding their composition, structure, and properties. Zhang\cite{zhang2016novel} employed naive bayes algorithm to classify mutagen and non-mutagens based on Descriptors and Extended Connectivity Fingerprints (ECFP-14). Likewise,
Zhang\cite{zhang2019lightgbm} conducted an investigation into the performance of the LightGBM algorithm for classifying compounds across various toxicological endpoints. The study employed RDKit descriptors and Morgan fingerprints as features to train the LightGBM model. The achieved area under the curve (AUC) values were $0.786$ and $0.8$ for RDKit descriptors and Morgan fingerprints, respectively. Advancements in deep learning have greatly influenced chemoinformatics applications, especially molecular property, affinity and interaction prediction \cite{@articlechen2018rise}. DeepTox\cite{mayr2016deeptox} calculates numerous chemical descriptors which are fed into ML techniques to predict toxicity. The process involves training multiple models, evaluating their performance, and subsequently assembling the best-performing models into ensembles.\\
\indent In the realm of Molecular Property Prediction (MPP), the significance of neural networks (NNs) lies in their capacity to discern intricate non-linear relationships. This is particularly crucial where molecular structures and features exhibit complex relations that may pose challenges for linear models to capture\cite{li2022deep}. By using deep architectures and embedding layers, NNs can automatically extract features and representations from unprocessed molecular data like SMILES, molecular images and molecular graphs, eliminating the need of manual feature engineering\cite{yi2022graph}. Winter\cite{winter2019learning} presented a model to extract fixed-size representation of equivalent semantics from variable-sized representation such as SMILES strings using a Recurrent Neural Network (RNN). Goa et al. proposed Chemception\cite{goh2017chemception}, a method that takes the 2D images of molecules for prediction of toxicity and other properties using Convolution  Neural Networks (CNNs). Graph Neural Networks (GNNs) have attained considerable significance in the field of cheminformatics recently, primarily due to their capacity to proficiently model and analyze graph-structured data. They capture the spatial relationships between atoms and bonds, enabling accurate predictions of molecular properties and activities. Li et al. introduced MutagenPred-GCNN\cite{li2021mutagenpred} for predicting mutagenicity. They use graph convolutional neural network (GCNN) and achieved AUC values of approximately $0.87\%$ and $0.83\%$. The model takes molecular graphs as input, learning molecular representations for making predictions. Karim\cite{karim2019efficient} employed a hybrid framework that combines Decision Trees (DT) and NNs. 
Despite significant efforts in using fingerprints for predicting mutagenicity, many fingerprints are still unexplored for this endpoint. Additionally, while achieving high accuracy is important, it is equally crucial to develop prediction models that integrate diverse features from multiple modalities and provide explainable predictions.  
In light of these considerations, our study introduces a methodology that uses multiple modalities which incorporates diverse information about chemical compounds: substructural, physicochemical, geometrical and topological. This is achieved through a combination of handcrafted features and a data-driven representation of molecular structures obtained using the Graph Attention Network (GAT). The primary contribution of our study include:
\begin{enumerate}
 \item We introduce a STacked Ensemble-based Mutagenicity prediction model (STEM) that uniquely integrates substructural, physicochemical, geometrical, and topological information about chemical compounds via multiple modalities.
\item To capture the structural information, we employ three fingerprints—KlekotaRoth, CDKExtended, and RDkit. Additionally, we derive 2D and 3D descriptors to account for physicochemical and geometric information, respectively. This approach represents the first use of KlekotaRoth and CDKExtended fingerprints in mutagenicity prediction.
\item For topological information, we utilize a Graph Attention Network (GAT), a novel integration into mutagenicity prediction. Our results demonstrate the effectiveness of GAT in learning representations that accurately predict the target variable.
\item We conducted an ablation study to highlight the contribution of different modalities in the overall model.
\item To interpret our method, we use SHapley Additive exPlanations (SHAP), which allows to understand the impact of different classifiers on the final prediction and to identify the most significant features.
\end{enumerate}
 
 \begin{table}[htbp!]
\centering
\caption{Statistical summary of the  datasets used}
\begin{tabular}{ c c c c } 
\label{tab:dataset summary}\\
 \hline
 \textbf{Dataset} & \textbf{Drugs} & \textbf{Non-mutagens} & \textbf{Mutagens}\\
 \hline
 Hansen et al.\cite{hansen2009benchmark} & 6277 & 2895 & 3382\\ 
 Xu et al.\cite{de2022toxcsm} & 8102 & 2347 & 3787 \\
 \hline
\end{tabular}
\end{table}
\section{Materials and Methods}
\subsection{Dataset description}
In this research, we utilized two datasets widely used in literature: the ames mutagenicity benchmark dataset by Hansen et al.\cite{hansen2009benchmark}, and the chemical ames mutagenicity dataset by Xu et al.\cite{xu2012silico}. The statistics are provided in Table \ref{tab:dataset summary}. The Hansen dataset documents the experimental outcomes of the Ames mutagenicity test for various compounds in the Chemical Carcinogenesis Research Information System (CCRIS), GeneTox, Helma, VITIC, and three other studies\cite{kazius2005derivation,helma2004data}. The dataset comprises 6512 compounds represented as canonical SMILES, along with the corresponding results of the Ames test. Among these, $3009$ compounds are identified as non-mutagens, while the rest are categorized as mutagens. We excluded duplicate molecules that shared identical canonical SMILES codes. In cases of conflicting Ames test outcomes, priority was given to positive results. Following this, we obtained a dataset comprising 6,277 molecules, with 3,382 categorized as the positive class and the remainder classified as the negative class. On the other hand, the Xu dataset\cite{de2022toxcsm} comprises $8102$ compounds with known mutagenicity properties. We removed $1968$ compounds that were common to the Hansen data. The rationale for eliminating common compounds is to establish a wholly distinct data to demonstrate the generalizability of our methodology across multiple datasets. 
The final dataset, thus comprises of $6134$ compounds, with $3787$ as mutagens and $2347$ as non-mutagens.
\subsection{Architecture of Proposed Methodology}
\begin{figure*}[ht]
    \centering
\includegraphics[width=1.0\textwidth]{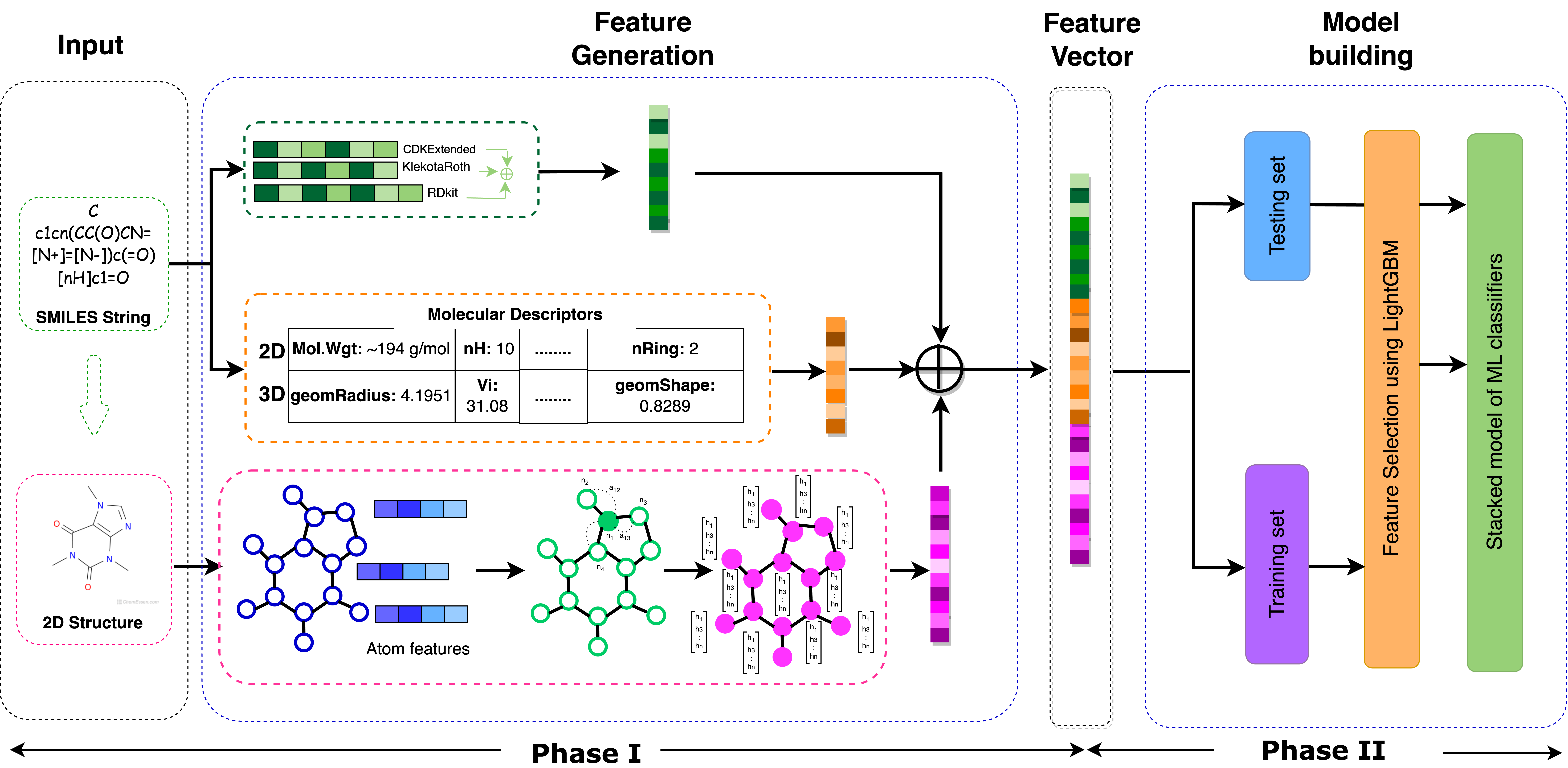}     
   \caption{The architecture of the proposed method STEM}
    \label{fig:flowdiagram}
\end{figure*}

\begin{figure*}[ht]
    \centering
\includegraphics[width=1.0\textwidth]{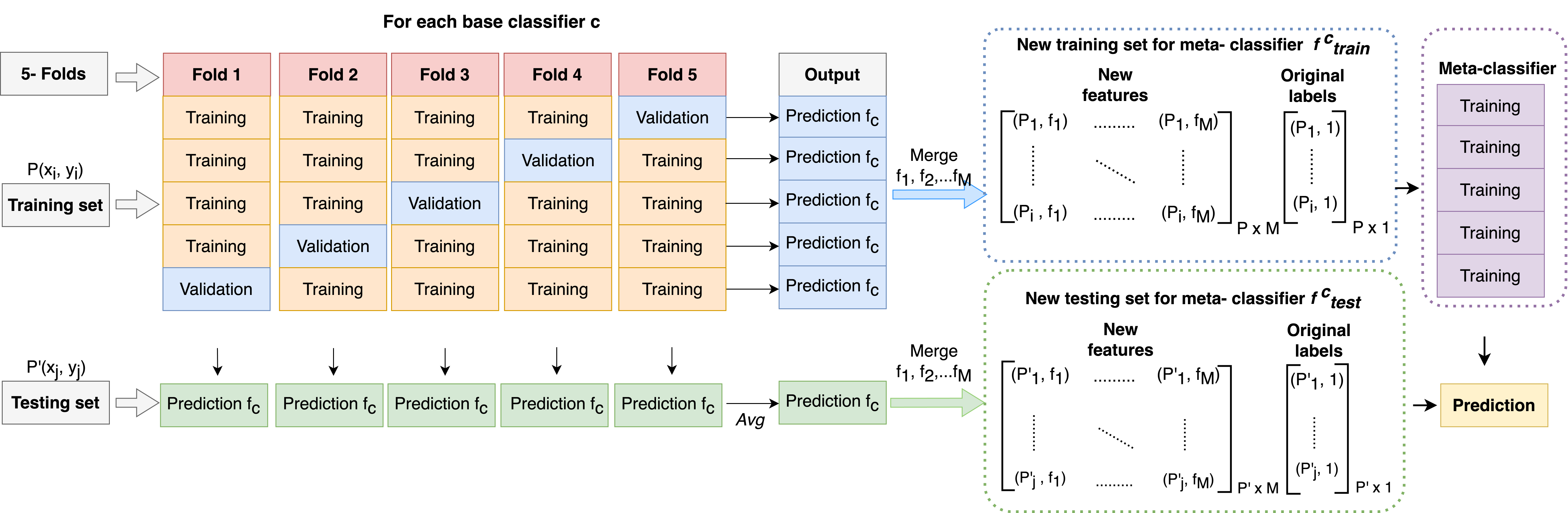}     
   \caption{The process of stacking with 5-fold CV}
    \label{fig:stacking5cv}
\end{figure*}

\begin{figure}
     \centering
     \begin{subfigure}[b]{0.45\textwidth}
         \centering
         \includegraphics[width=\textwidth]{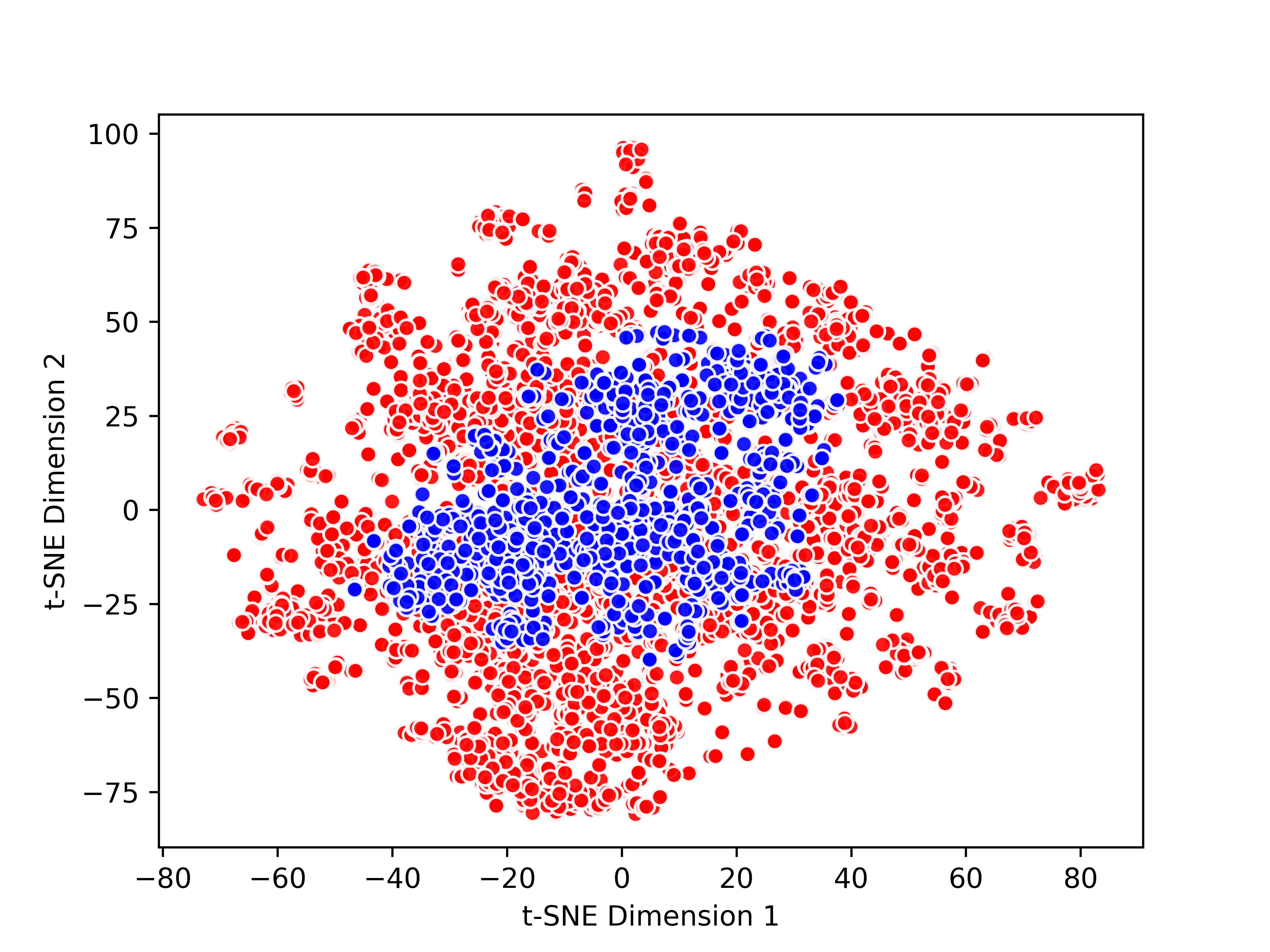}
         \caption{}
         \label{fig: embeddings_x}
     \end{subfigure}
     \begin{subfigure}[b]{0.45\textwidth}
         \centering
         \includegraphics[width=\textwidth]{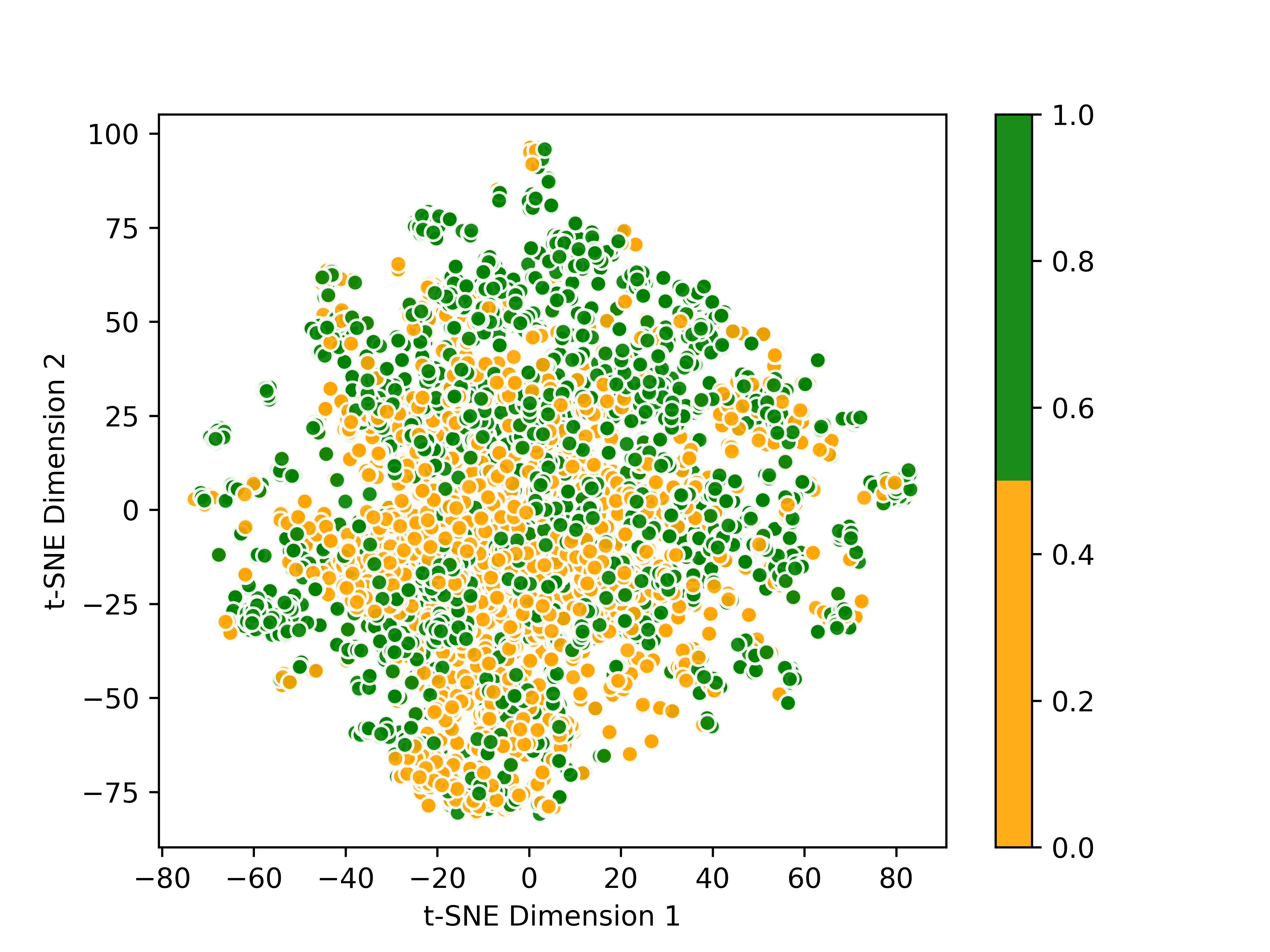}
         \caption{}
         \label{fig: embeddings_y}
     \end{subfigure}
        \caption{The 2D t-SNE representation of the gathered compounds illustrates: (A) the dataset encompassing features (B) associated labels. The dimensions t-SNE1 and t-SNE2 result from the reduction of the original feature space}
        \label{fig: two embeddings}
\end{figure}

In this paper, we propose a mutagenicity prediction model using multiple modalities with stacked ensemble of various machine learning classifiers. Our approach is illustrated in Figure \ref{fig:flowdiagram}. 
The proposed model is separated into two major phases: Phase I and Phase II. 
\begin{itemize}
    \item In \textbf{Phase I}, chemical molecules are encoded to yield diverse representations that encapsulate distinct information. We leverage molecular fingerprints, 2D and 3D descriptors, and graph-based molecular representations for this purpose. Molecular fingerprints and descriptors are derived from the SMILES notations of compounds. Graph-based molecular representations are generated from molecular graphs via Graph Attention Network (GAT). 
    The operations involve in our GAT model is illustrated in Figure \ref{fig: gat}. Details on the encoded representations is provided in the subsequent subsection. All the representations are integrated based on equation \ref{eq1}. 
    \begin{equation}
    \label{eq1}
        V = (R_{fp} || R_{desc} || R_{gat} )
    \end{equation}
    
Here, $R_{desc}$ denotes the vector encompassing all descriptors, of which 772 are 1D and 2D descriptors, while 215 are 3D descriptors. $R_{gat}$ represents the encoded representation from GAT,  and $R_{fp}$ is the concatenated vector of all three fingerprints as shown in equation \ref{eq2}.
    \begin{equation}
    \label{eq2}
        R_{fp} = (FP_{CDK} || FP_{RDKit} || FP_{Klekotaroth})
    \end{equation}
The resultant vector thus formed, serves as input for Phase II.

 \begin{figure*}[ht!]
    \centering
\includegraphics[width=1.0\textwidth]{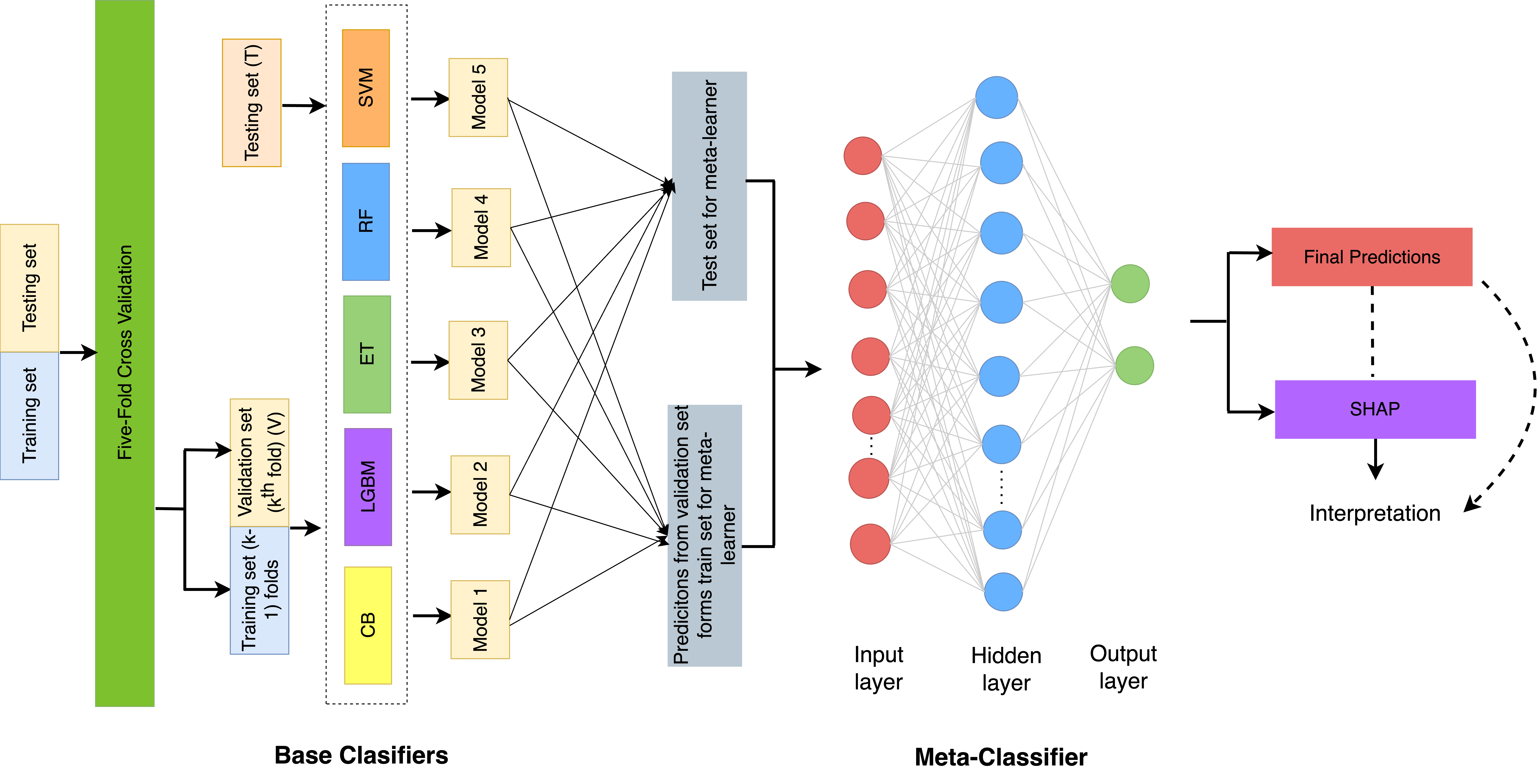}     
   \caption{The architecture of  the stacked model of ML classifier}
    \label{fig:stacked model}
\end{figure*}
\item  In \textbf{Phase II}, the data is partitioned into training and testing set in a 80:20 ratio. We employ t-SNE analysis to study the distribution of data points in a low dimensional space as shown in Figure \ref{fig: two embeddings}. The apparent overlap of the training and testing datasets (Fig. \ref{fig: embeddings_x}), as well as the overlap of chemicals with different mutagenicity labels (Fig. \ref{fig: embeddings_y}) which demonstrates that there is structural variation within the dataset and the division into training and testing sets is reasonable. To address redundancy and high-dimensionality issues arising from combining various representations, we use LightGBM \cite{ke2017lightgbm}. It efficiently computes feature significance scores, capturing non-linear correlations in the data within large datasets, which makes it a preferred choice for feature selection. The derived vector serves as input for the stacked model shown in Figure \ref{fig:stacked model}. First, we use classifiers such as Support Vector Machine (SVM)\cite{cortes1995support}, Random Forest (RF)\cite{breiman2001random}, LightGBM (LGBM)\cite{ke2017lightgbm}, Catboost (CB)\cite{prokhorenkova2018catboost}, and ExtraTrees (ET)\cite{geurts2006extremely} to forecast the mutagenicity of compounds. The predictions generated by these classifiers on the validation sets are then stacked and utilized as new features for training a neural network. The neural network is constructed to understand the complex non-linear associations between the predictions made by different classifiers and the target variable. We perform hyperparameter tuning using a random search approach. Detailed information about the hyperparameter settings can be found in Table \ref{tab:hyper summary}. All the experiments are conducted using 5-fold cross validation (CV) method. Figure \ref{fig:stacking5cv} illustrates the process of stacking with 5-fold CV. Given a training data of size $P$ with features $x_i$ and labels $y_i$, and a testing data of size $P'$ with features $x_j$ and labels $y_j$, the following steps were used to implement stacking: 
  \begin{enumerate}
        \item The training set is divided into five equal folds. For each base classifier $c$, four folds are used for training, and the remaining fold is used for validation. This process is repeated five times for 5-fold CV. 
        All the predictions of model $c$ from each fold form a new training feature set $f_{train}^c$ for the stacked model. 
        Simultaneously, the average of the prediction results on the testing set is computed for  classifier $c$ across each fold. This forms a new feature set $f_{test}^c$ for testing on the stacked model
        \item Repeat the above step on all the classifiers, forming a matrix of size $P\times M$. The test data results in a matrix of size $P'\times M$. Retain the original labels of this dataset as the labels for these new features. Here, 
        $M$ is the no. of new features. 
        In our case, its value is 5. 
        \item Train the DNN on the new features and test it on the test data obtained from step 2.
        
    \end{enumerate}
    
\end{itemize}

\begin{table}
\centering
\caption{Description about the hyperparameter settings}
\scalebox{0.55}{
\begin{tabular}{ c c } 
\label{tab:hyper summary}\\
 \hline
 \textbf{Models} & \textbf{Hyperparameter values} \\
 \hline\\
 LightGBM & \makecell{‘objective’: binary,  ‘metric’: binary\_logloss, 
   ‘boosting\_type’: gbdt, \\ 
 ‘num\_leaves’: 31,   ‘learning\_rate’: 0.1,   ‘feature\_fraction’: 0.9, 
  ‘n\_estimators’: 200}  \\ \\
 ExtraTrees & \makecell{‘n\_estimators’: 350, ‘criterion’: ‘gini’, ‘min\_samples\_split’: 2, ‘min\_samples\_leaf’: 1 }  \\\\
 Random Forest & \makecell{‘n\_estimators’: 120, 
 ‘max\_depth’: 17, ‘min\_samples\_split’: 2, ‘min\_samples\_leaf’: 1}  \\\\
 Support Vector Machines & \makecell{‘kernel’: linear, ‘C’: 1.0} \\\\
 CatBoost & \makecell{‘iterations’: 40, ‘learning\_rate’: 0.2} \\\\
GAT &  \makecell{‘iterations’: 40, ‘learning\_rate’: 0.2, ‘no. of attentions heads’: 8, ‘hidden\_size’: 300, ‘batch\_size’: 32, drop\_out = 0.2 }\\\\
DNN & \makecell{‘no. of hidden\_layers’: 1, ‘learning\_rate’: 0.001, ‘hidden\_neurons’: 100, ‘batch\_size’: auto, ‘hidden\_layer\_activation’: relu, ‘solver’: adam}\\\\
 \hline
\end{tabular}}
\end{table}
\subsection{Feature Representation}
The dataset provides SMILES strings representing the chemical structures of both mutagenic and non-mutagenic compounds. SMILES represents chemical structures using ASCII characters in a linear form. These SMILES are fed as input to encode various numerical representations.

\subsubsection{Substructural features}
We represent the substructural details of chemical compounds using three molecular fingerprints: KlekotaRoth, RDKit, and CDKExtended. To compute these fingerprints, we use the tool padelpy\cite{kesslerpadelpy}. In total, we generated fingerprint vectors  corresponding to 1024D RDKit fragments, 4860D KlekotaRoth fragments, and 1024D CDKExtended fragments. The binary vectors represent the existence or non-existence of particular substructures, where each bit functions as a feature during the learning process.. 

\subsubsection{Physicochemical features}
We calculated 1D, 2D, and 3D molecular descriptors to capture physicochemical and geometric aspects, providing a fundamental understanding of molecule structure and properties. For instance, basic features like molecular weight, molar refractivity, atom and bond counts are captured by 1D descriptors,
the connection of atoms in the molecular plane is taken into consideration by 2D descriptors and
3D descriptors take into account the three-dimensional arrangement of atoms to provide details about the geometry of molecules. 
The mordred library\cite{moriwaki2018mordred} is used to conduct feature extraction. We only included descriptors with low variance ($\leq 0.05$) and eliminated those with NULL values to guarantee data consistency and quality. Using min-max scaling, the resulting feature is normalized to a range of 0 to 1.
\subsubsection{Graph-based Spatial features}
While 3D descriptors also capture some spatial information, using GNNs to capture the spatial interactions between atoms and bonds in a molecule is beneficial in many ways. Variable bond lengths, angles, and torsional angles are examples of the variations that are frequently seen in molecular structures. Since GNNs work on the graph structure, they capture a variety of molecule geometries with flexibility, making them well-suited to manage such variations. Conventional descriptors frequently result in a considerable loss of molecular information by condensing it into a predetermined set of features. In contrast, GNNs preserve a more flexible and expressive representation of chemical structures \cite{scarselli2008graph}. The attention mechanism in GNNs provides interpretability by indicating which neighbors contribute the most to each node's representation \cite{cai2022fp}. We therefore use a GAT framework to encode a hierarchical representation of a molecular graph that represents spatial information preserving local and global features through the recurrent aggregation of data from nearby atoms. 
\subsection{Models}
\subsubsection{Graph Attention Network}

A molecule can be expressed as a graph G = (V, E), with V representing the atoms in the molecule and E denoting the bonds connecting these atoms. Each molecular graph is represented by two matrices serving as features. The first matrix is the adjacency matrix  $A_{N\times N}$ which encapsulates the connectivity information. 
It provides a structured representation of the graph's topology, indicating which atoms are connected to each other. The second matrix is the node feature matrix $F \in R^{{N\times X}}$, where ‘N’ are the atomic nodes and ‘X’ are the atomic attributes with each node. The combination of these matrices allows the GNN to learn and process both the structural links between atoms and the individual attributes associated with each atom in the molecular graph. The atomic attributes we used in our work are given in Supplementary materials. Message passing is the central operation in GNNs where each node gathers information from its neighbors updating its node representation based on the features of surrounding nodes. Mathematically, given a set of node features $h_v$ for node $v$ and $h_u$ for all nodes $u \in neighbor(v)$, this can be expressed as Equation \ref{aggregate}.
\begin{equation}
\label{aggregate}
    h_{v}^{'} = \sigma\left(h_{v}\sum\limits_{u\in neighbor(v)}^{} h_u\right)
\end{equation} 
where $\sigma$ is the aggregation function. The entire graph representation is obtained by aggregating the updated representations of all its atoms. 
\begin{equation}
    \label{sum}
    H = \sigma\left(\sum\limits_{v\in G} h_{v}^{'}\right)
\end{equation}
To enhance the message-passing process, GAT\cite{velickovic2017graph} uses an attention mechanism that assigns different weights to the neighbors of the node during information aggregation. This allows to selectively focus on different neighbors. The attention coefficient $\phi_{uv}$ between a node $v$ and $u$ is computed as Equation \ref{attention} and \ref{softmax}. 

\begin{equation}
    \label{attention}
    \phi_{uv} = LeakyReLU\left(a^{T}\left(\left[ W_{1}h_{v}|| W_{1}h_{u}\right]\right)\right) 
\end{equation}
where $||$ denotes the concatenation, $W$  is the learnable weight matrix, $a$ is the learnable attention vector, $LeakyReLU$ is the Linear Rectified Units activation function.  To normalize the  unnormalized attention scores, we use a softmax function as shown below in Equation \ref{softmax}.
\begin{equation}
     \label{softmax}
    \alpha_{uv} = \frac {exp\left(\phi_{uv}\right)}{\sum\limits_{k\in neighbor(v)}exp \left(\phi_{vk}\right)}
\end{equation}

\begin{figure}
     \centering
         \includegraphics[width= 1.1\textwidth]{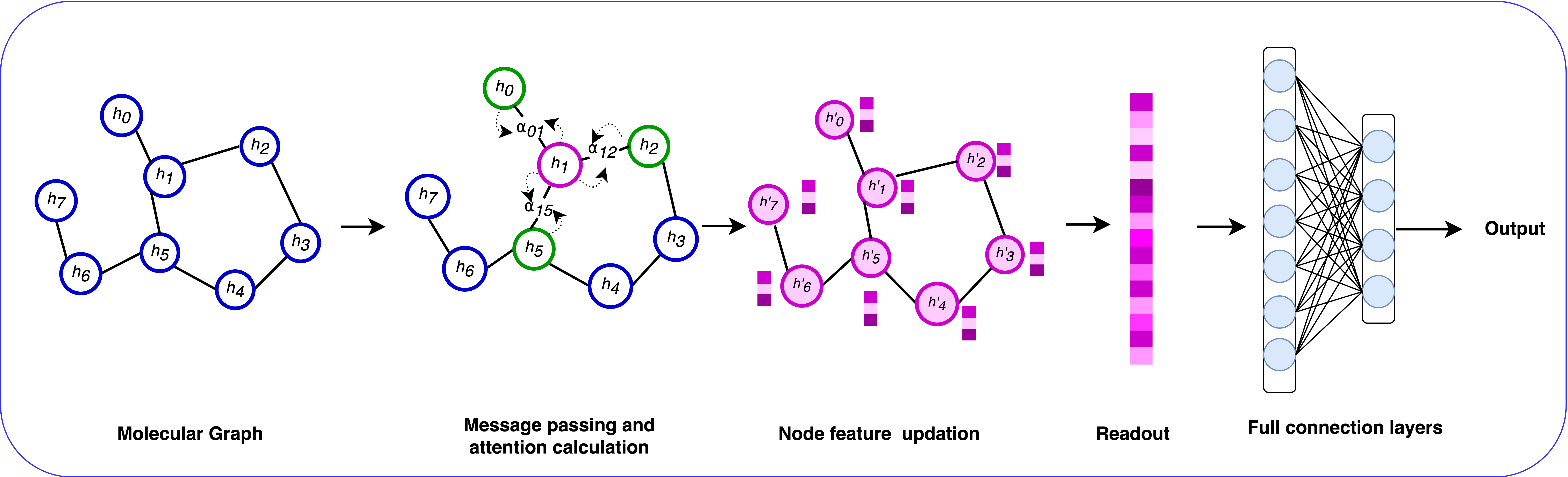}
         \caption{The flow diagram of Graph Attention Network}
         \label{fig: gat}
\end{figure}

 To obtained the final node representations, a weighted sum of all the neighbors node features are taken as shown in Equation \ref{final}.

\begin{equation}
    \label{final}
    h_{v}^{'} = \sigma \left(\sum\limits_{u\in neighbor(v)} \alpha_{uv} W \cdot h_{u}\right)
\end{equation}
We repeatedly computed attention coefficients multiple times and calculated the average as the final attention.
After updating all the nodes,  the final representation of the molecular graph is the mean of all its node representations illustrated in Equation \ref{graph}. The framework for graph attention network is as depicted in Figure \ref{fig: gat}.
\begin{equation}
\label{graph}
    G = \frac{1}{N} \sum\limits_{v\in G} h_{v}^{'}
\end{equation}

\subsubsection{Support Vector Machine}

Support Vector Machines (SVMs) are supervised learning algorithms employed for tasks involving classification and regression. They identify a hyperplane within a high-dimensional space that maximizes the margin between two classes. The location of the hyperplane is determined by support vectors, which are the data points situated in close proximity to the decision boundary. Through a kernel approach, SVMs can manage non-linear boundaries by translating features into a higher-dimensional space \cite{huang2018applications}. Using a random search, we tuned two crucial SVM hyperparameters, regularization C and kernel function. For linear separability, we employ the linear kernel function given below:
\begin{equation}
    L(x,x_i) = x \cdot x_i
\end{equation}
where x and $x_i$ are the input and the $i^{th}$ support vector, respectively.

\subsubsection{Random Forest}

Random Forest is an ensemble learning method that leverages the power of multiple decision trees to enhance predictive accuracy and robustness. Each decision tree is trained on a randomly selected subset of the data using a technique called bootstrap sampling, where subsets are selected with replacement. Additionally, at each node of a tree, a random subset of features is considered for splitting, promoting diversity among the trees and preventing overfitting \cite{rigatti2017random}. The RF hyperparameter  described as the number of trees in the forest (n\_estimators) and maximum depth (max\_depth) are important dynamic parameters that we set to 120 and 17 respectively.

\subsubsection{LightGBM}
Though Random forest and LightGBM both are tree based approaches, but due to its inherent computational complexity, RF performance might be impacted when dealing with extremely large datasets with high-dimensional feature spaces. LGBM's efficient leaf-wise growth and histogram-based binning make it particularly well-suited for handling such datasets \cite{ke2017lightgbm}. It can perform well on large-scale problems and introduces features like early stopping, which halts training when performance plateaus, and categorical feature support, enhancing its applicability to various datasets.

\subsubsection{Extra Trees}
Extra Trees, or Extremely Randomized Trees, is also an ensemble learning method similar to Random Forests.  It constructs numerous decision trees through bootstrap sampling and employs a random subset of features at each split. However, Extra Trees goes a step further by selecting random thresholds for feature splits, making it even more robust to overfitting. This randomness in both feature selection and threshold choice enhances diversity among trees. The additional layer of randomness can make them robust to handle outliers and noise in the dataset better than certain gradient boosting methods\cite{geurts2006extremely}.

\subsubsection{CatBoost}
Developed by Yandex, CatBoost efficiently handles categorical variables without the need for extensive preprocessing. The algorithm uses a depth-aware strategy for tree growth, improving predictive accuracy \cite{prokhorenkova2018catboost}. With robust support for parallelization, CatBoost is computationally efficient. Its strength lie in its ability to deliver competitive performance with minimal hyperparameter tuning, making it an effective choice for a wide range of classification and regression tasks.
\\\\
We stacked these models to complement each other. For instance, SVM capture complex decision boundaries, RF and Extra Trees provide robustness, LGBM deals with high-dimensional features space, and CatBoost handles categorical features efficiently. Combining these strengths can lead to an ensemble with improved stability and enhanced robustness capturing both linear and nonlinear patterns in the data.
\subsubsection{Deep Neural Network}
We use a DNN\cite{anderson1995introduction} that takes as input the predictions of all the ML classifier and generate output. The network consist of an input layer, an output layer and a single hidden layer. The dimensions of the hidden layer and other hyperparameters are stated in Table \ref{tab:hyper summary}. We use ReLU activation function to expedite the convergence and prevent overfitting. Given $X = \{p_1, p_2, \cdots p_n\}$ as the predictions of the ML model, the output of the hidden layer is computed as:
\begin{equation}
\label{hidden}
    Z_{1} = W_{1} \cdot X + b_{1}
\end{equation}

\begin{equation}
\label{hidden_activation}
        a_{1} = \sigma\left(Z_{1}\right)
\end{equation}

The output layer calculations are:
\begin{equation}
\label{output}
    Z_{2} = W_{2} \cdot a_{1} + b_{2}
\end{equation}

\begin{equation}
\label{output_activation}
        a_{2} = \varphi\left(Z_{2}\right)
\end{equation}

where $\sigma$ and $\varphi$ are the ReLU and sigmoid activation function, $W_{1}$, $b_{1}$ and $W_{2}$, $b_{2}$ are the weights and biases of the hidden layer and the output layer respectively.
The final layer of the neural network is activated by a sigmoid function, and its output is optimized using the Binary Cross-Entropy (BCE) loss function. To expedite the optimization process for deep neural networks, the ADAM optimizer is employed, known for its faster convergence compared to traditional optimization algorithms\cite{kingma2014adam}. This choice is made to enhance the efficiency of the training process and ensure effective learning of the model parameters.


\begin{table}[htbp!]
\centering
\caption{Various evaluation metrics used in this study. TP (true positives) denotes successfully identified mutagens, TN (true negatives) denotes successfully identified non-mutagens, FP (false positive) denotes wrongly identified mutagens, FN (false negatives) denotes wrongly identified non-mutagens.}
\scalebox{0.75}{
\begin{tabular}{c c p{10cm}} \hline
\label{tab: metrics}

 \textbf{Metric} & \textbf{Expression} & \textbf{Description} \\\hline
 \\
 Accuracy & \(\displaystyle \frac{TP+TN}{TP+FP+TN+FN} \) & Percentage of correctly classified samples representing an overall measure of the model’s correctness\\ 
 Precision & \(\displaystyle \frac{TP+FP}{TP} \) & A measure of the accuracy of positive predictions, calculated as the ratio of correctly predicted positive observations to the total predicted positives. \\
 Recall & \(\displaystyle \frac{TP+FN}{TP} \) & A ratio of correctly predicted positive observations to all the actual positives. Recall measures the model ability to accurately capture all positive instances. \\
 F1-Score & \(\displaystyle 2 * \frac{Precision * Recall}{Precision + Recall} \) & The harmonic mean of precision and recall,  capturing a balanced measure of a model's performance. \\
 AUC & \(\int_{0}^{1} TPR(FPR^{-1}(t)) \,dt\) & AUC (Area Under the Curve) signifies the area under the Receiver Operating Characteristic (ROC) curve. The ROC curve illustrates the true positive rate ($TPR$) against the false positive rate ($FPR$) at different thresholds ($t$). AUC serves as a metric for evaluating model capacity to differentiate between classes. A higher AUC-ROC value, closer to 1, indicates superior discriminatory performance.\\
 AUPR & \(\int_{0}^{1} Precision(Recall^{-1}(t)) \,dt\) & AUC-PR denotes the Area Under the Precision-Recall curve. This curve illustrates precision plotted against recall at different thresholds. A higher AUC-PR value (closer to 1) indicates a model that maintains high precision across various recall levels. \\ \hline
 
\end{tabular}}
\end{table}
\section{Results }
\subsection{Performance evaluation}
We evaluated the classification effectiveness of our proposed strategy using six performance evaluators, as shown in Table \ref{tab: metrics}.
To avoid potential biases and overfitting, we use a 5-fold CV approach to evaluate the model performance. The training set was divided into five equal portions at random, with four parts used for model training and the remaining part saved for model evaluation. To reduce the impact of random fluctuations and to quantify the variability and uncertainty in the results, we repeated this process ten times on different seeds. The values thus obtained were averaged to evaluate the overall performance. 
\subsection{Performance of the proposed method STEM}
\begin{figure}
     \centering
     \begin{subfigure}[b]{0.45\textwidth}
         \centering
         \includegraphics[width=\textwidth]{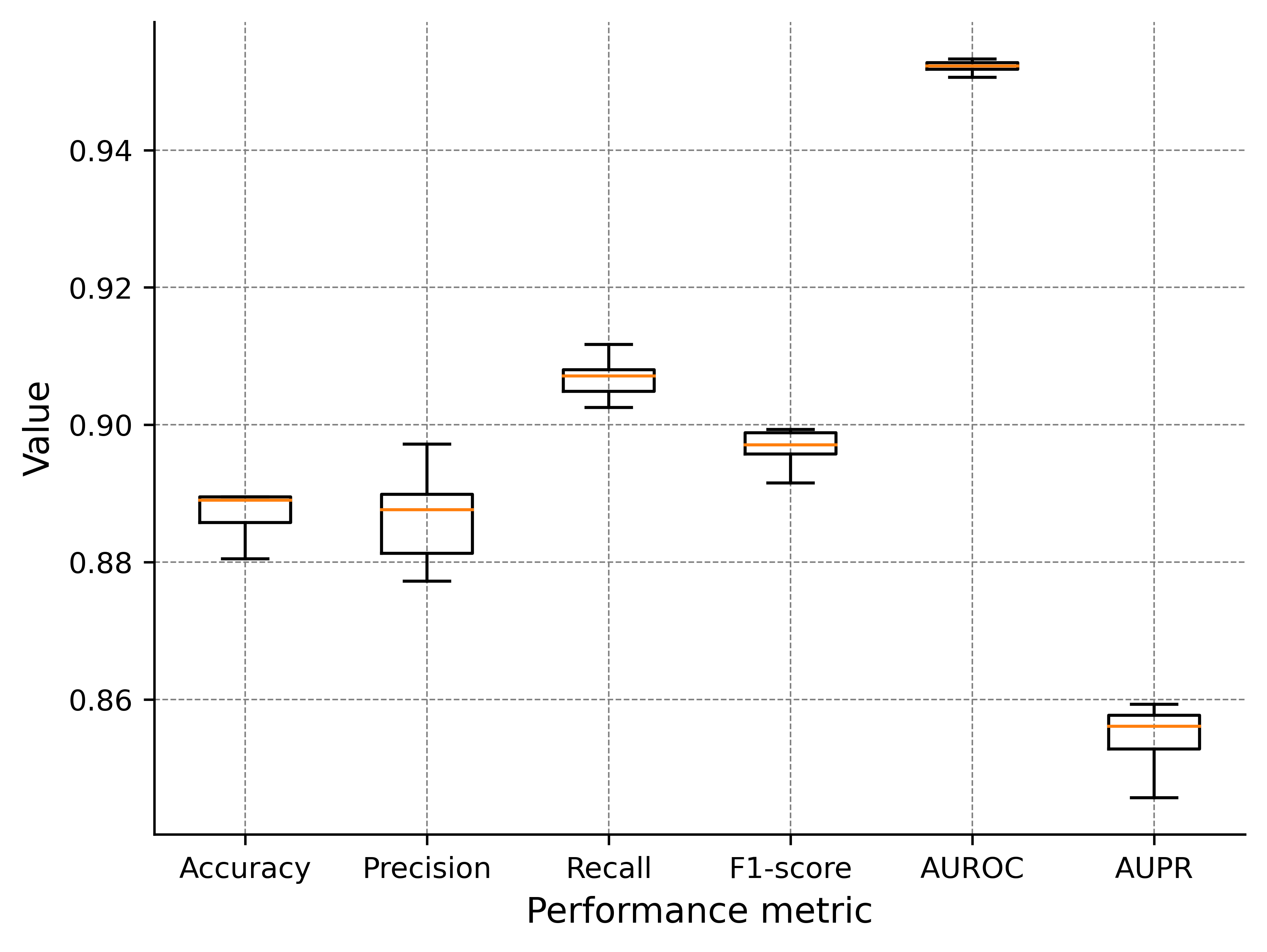}
         \caption{}
         \label{fig: seedeffect_test}
     \end{subfigure}
     \begin{subfigure}[b]{0.45\textwidth}
         \centering
         \includegraphics[width=\textwidth]{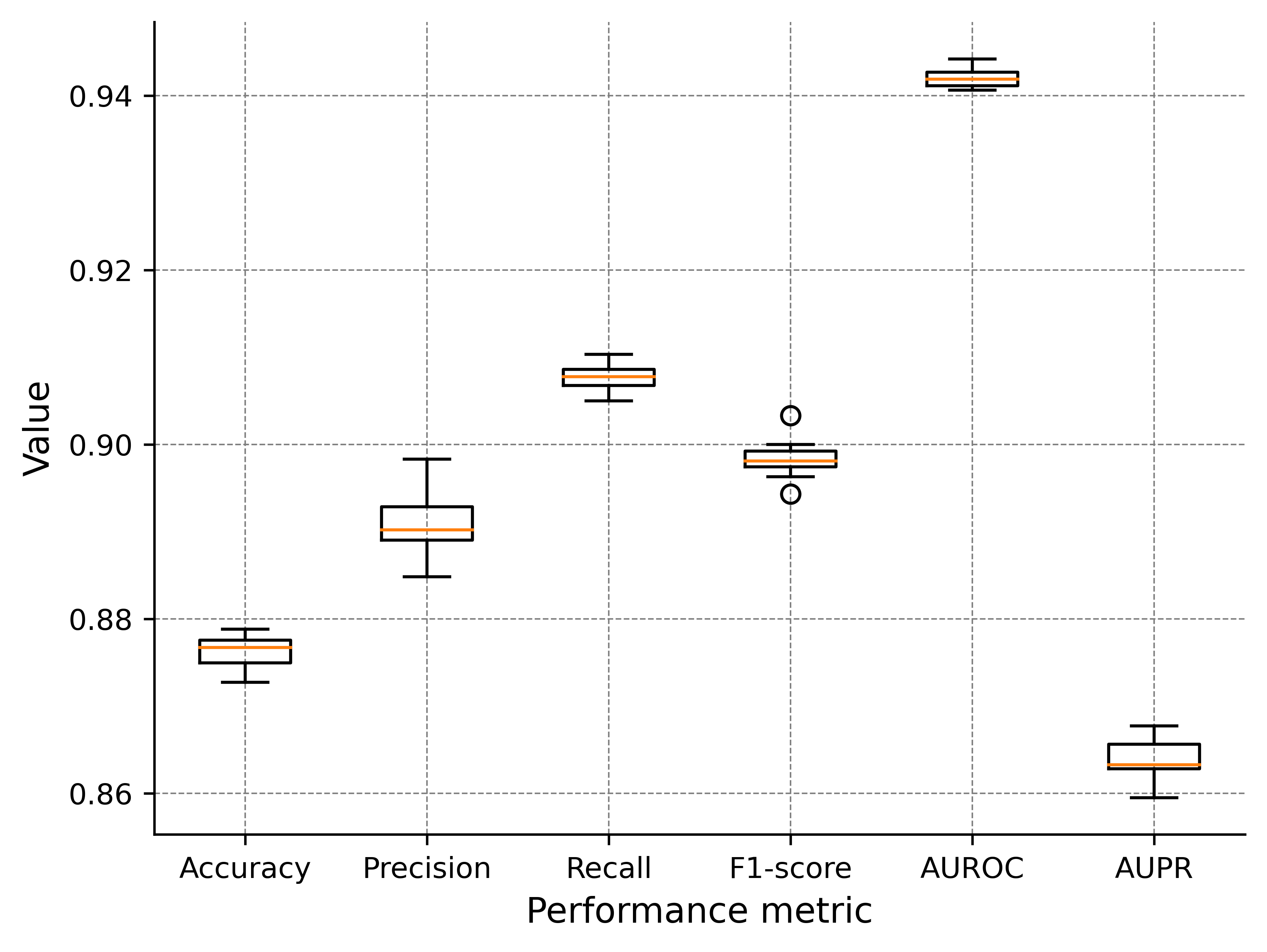}
          \caption{}
         \label{fig: seedeffect_valid}
     \end{subfigure}
     
        \caption{The change in metric results across 10 random seeds  during blind test on both (a)  Hansen and  (b) Xu dataset.}
        \label{fig:seed_effect}
\end{figure}

\begin{figure}[htbp!]
     \centering
     \begin{subfigure}[b]{0.49\textwidth}
         \centering
         \includegraphics[width=\textwidth]{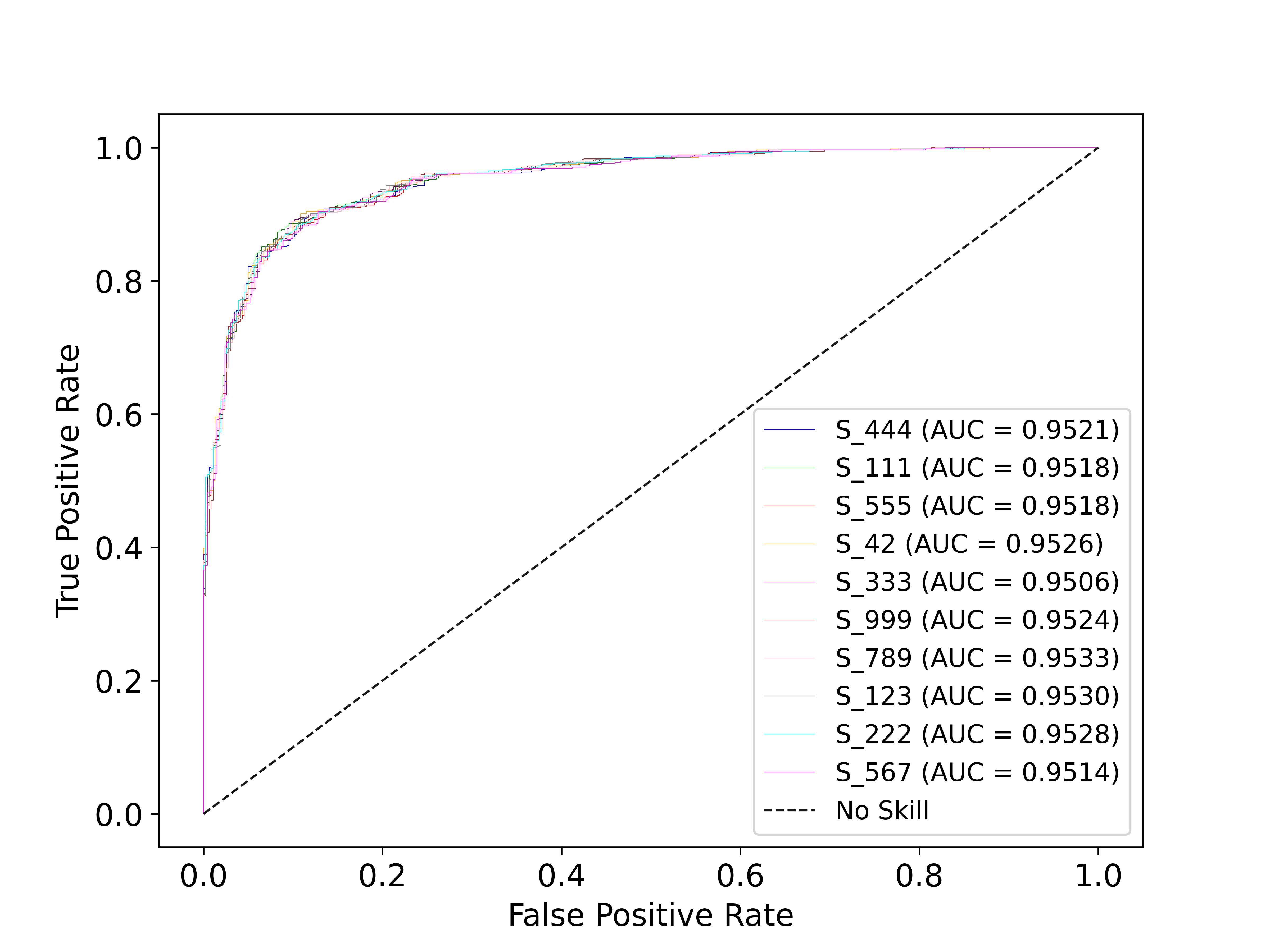}
         \caption{}
         \label{fig:ROC_test}
     \end{subfigure}  
     \begin{subfigure}[b]{0.49\textwidth}
         \centering
         \includegraphics[width=\textwidth]{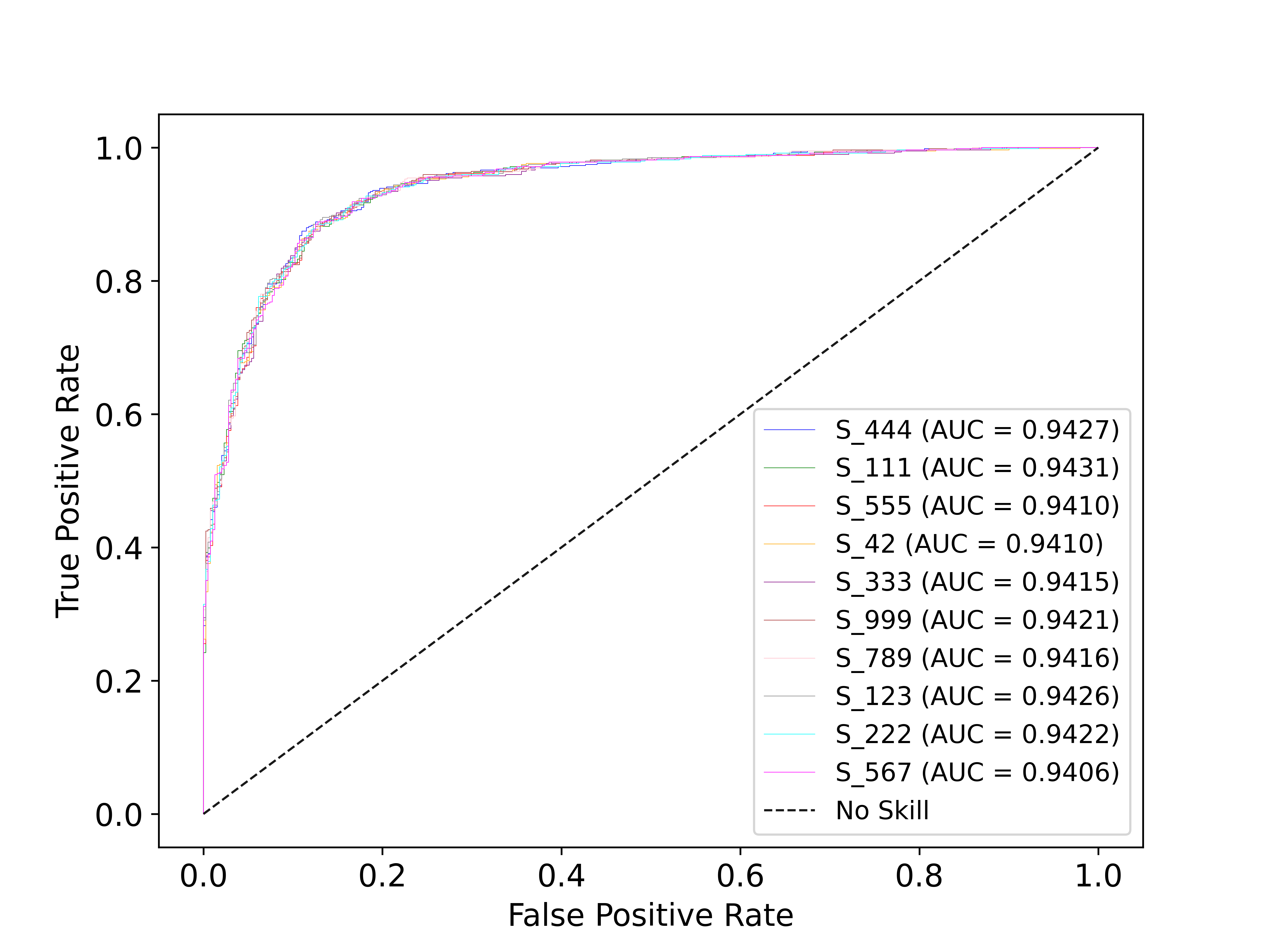}
          \caption{}
         \label{fig: ROC_train}
     \end{subfigure}
        \caption{The ROC curves of the models across all seeds are analyzed during blind test on both (a)  Hansen and  (b) Xu dataset.}
        \label{fig:ROC curves}
\end{figure}

In this research, we demonstrate that amalgamating diverse information regarding molecules yields a more intricate representation. This nuanced representation significantly enhances the accuracy of machine learning models in predicting mutagenicity. 
\begin{table}[htbp]
    \centering
    \caption{Performance evaluation of STEM using various metrics on the Hansen test set}
    \scalebox{0.85}{
    \begin{tabular}{p{1.8cm} p{1.8cm}p{1.8cm}p{1.8cm}p{1.8cm}p{1.8cm}p{1.8cm}} \\\hline
         \textbf{Seed} & \textbf{Accuracy} & \textbf{Precision} & \textbf{Recall} & \textbf{F1-score} & \textbf{AUC} & \textbf{AUPR}  \\\hline\\
        42 & 0.8855 & 0.8837 & 0.9080 & 0.8957 & 0.9533 & 0.8523\\
        123 & 0.8825 & 0.879 & 0.9080 & 0.8933 & 0.9526 & 0.8479\\
        567 & 0.8895 & 0.8929 & 0.9044 & 0.8986 & 0.9505 & 0.8593\\
        789 & 0.8895 & 0.89 & 0.9080 & 0.8989 & 0.9521 & 0.858\\
        999 & 0.8895 & 0.8872 & 0.9117 & 0.8993 & 0.9518 & 0.8567\\
        111 & 0.8865 & 0.8866 & 0.9062 & 0.8963 & 0.9527 & 0.8543 \\
        222 & 0.8885 & 0.8805 & 0.9044 & 0.8978 & 0.953 & 0.8578\\
        333 & 0.8895 & 0.8886 & 0.9099 & 0.8991 & 0.9523 & 0.8573\\
        444 & 0.8805 & 0.8772 & 0.9062 & 0.8915 & 0.9517 & 0.8457\\
        555 & 0.8865 & 0.8894 & 0.9025 & 0.8959 & 0.9514 & 0.8555 \\
        \hline
        \textbf{Avg} & \textbf{0.8868 \textpm 0.003} & \textbf{0.8855 \textpm 0.006} & \textbf{0.907 \textpm 0.002} & \textbf{0.8966 \textpm 0.002} & \textbf{0.9521 \textpm0.001} & \textbf{0.8544 \textpm 0.004}\\\hline
        
    \end{tabular}}
    \label{tab:test_all_seeds}
\end{table}
To verify the robustness and generalizability, the final model STEM is validated against a test set. We showed that combining substructural, physicochemical, geometrical and graph-based spatial features elevates the performance beyond present state-of-the-art methods achieving a peak AUC of about 0.95 on the Hansen benchmark dataset. For the blind test sets of the two datasets, the comprehensive metric findings for all 10 random seeds are reported in Table
\ref{tab:test_all_seeds} and Table \ref{tab:toxcsm_all_seeds} respectively. We obtained 88.68\%  accuracy, 88.55\% precision, 90.7\% recall, 89.66\% f1-score, 95.21\% AUC and 85.44\% AUPR on the Hansen dataset. Likewise, an accuracy, and AUC of 87.65\%, and 94.2\% is achieved on the Xu dataset. The corresponding AUC-ROC curve for both the Hansen and Xu dataset is visualized in Figure \ref{fig:ROC curves}.
The observation depicted in Figure \ref{fig:seed_effect} reveals a remarkable stability in key performance metrics across diverse random seeds, underscoring the reliability of the model. Notably, metrics such as ROC-AUC, accuracy, and F1-score exhibit minimal variability with short box length  across both the datasets.
\begin{table}[htbp]
    \centering
    \caption{Performance evaluation of STEM using various metrics on the Xu \cite{xu2012silico} test set}
    \scalebox{0.85}{
    \begin{tabular}{p{1.8cm} p{1.8cm} p{1.8cm} p{1.8cm} p{1.8cm} p{1.8cm} p{1.8cm}} \\\hline
         \textbf{Seed} & \textbf{Accuracy} & \textbf{Precision} & \textbf{Recall} & \textbf{F1-score} & \textbf{AUC} & \textbf{AUPR}  \\\hline\\
        42 & 0.8778 & 0.8905 & 0.9086 & 0.8994 & 0.9442 & 0.8641\\
        123 & 0.8727 & 0.8935 & 0.9061 & 0.8943 & 0.9415 & 0.863\\
        567 & 0.8768 & 0.8983 & 0.9067 & 0.8975 & 0.9421 & 0.8677\\
        789 & 0.8788 & 0.8933 & 0.9069 & 0.9 & 0.9416 & 0.8662\\
        999 & 0.8767 & 0.889 & 0.9086 & 0.8987 & 0.941 & 0.8628\\
        111 & 0.8767 & 0.889 & 0.9086 & 0.8987 & 0.9426 & 0.8628 \\
        222 & 0.8747 & 0.8848 & 0.9103 & 0.8974 & 0.9431 & 0.8595\\
        333 & 0.8737 & 0.8859 & 0.9069 & 0.8963 & 0.9406 & 0.8595\\
        444 & 0.8757 & 0.8914 & 0.9035 & 0.8974 & 0.941 & 0.8635\\
        555 & 0.8818 & 0.8899 & 0.917 & 0.9033 & 0.9427 & 0.8661 \\
        \hline
        \textbf{Avg} & \textbf{0.8765 \textpm 0.002} & \textbf{0.8905 \textpm 0.003} & \textbf{0.9062 \textpm 0.003} & \textbf{0.8983 \textpm 0.002} & \textbf{0.942 \textpm 0.001} & \textbf{0.8635 \textpm 0.002}\\\hline
        
    \end{tabular}}
    \label{tab:toxcsm_all_seeds}
\end{table}
This consistent performance across different seed initializations implies that the model's predictive capacity remains robust and unaffected by varying starting conditions. Furthermore, while precision and recall metrics display slight variations, their consistency across both sets suggest a balanced model that avoids overfitting. The marginal fluctuations observed in these metrics, coupled with their persistence across random seeds, reinforce the model's stability and indicate that it generalizes well to unseen test data.
\subsection{Comparison against state-of-the-art methods}
Several publications have appeared in previous years that use machine learning techniques to predict mutagenicity. Table \ref{tab:sota} shows some computational methods for predicting the mutagenicity of compounds on the Hansen benchmark dataset.  To provide a fair and equitable comparison, we used the same evaluation criteria as these approaches. The stacking model developed in our study outperforms these methods in terms of AUC.

As shown in table \ref{tab:sota}, STEM achieves an average AUC of 0.9521, which is 11.21\%, 9.21\%, 7.41\%, 2.21\% higher than Zhang et al.\cite{zhang2017novel}, Hansen et al.\cite{hansen2009benchmark},  {MutagenPred-GCNN}\cite{li2021mutagenpred}, and 
Shinada et al.\cite{shinada2022optimizing} respectively. This increased AUC suggests that our probability forecasts are more accurate, making the results more valuable. The comparison with previously published models demonstrates that the strategy used in this work outperforms others in predicting the mutagenicity of chemical compounds.
\begin{table}[htbp]
    \centering
    \caption{Comparison of STEM  with state-of-the-art (SOTA) approaches using the Hansen et al. benchmark dataset for in silico prediction of Ames mutagenicity}
    \scalebox{0.7}{
    \begin{tabular}{c p{3.5cm} c c p{3.0cm} c } \\\hline
         \textbf{Method} & \textbf{Input} & \textbf{5-CV} & \textbf{Randomization} & \textbf{Algorithm} & \textbf{AUC}  \\\hline
        \textbf{Zhang et al.}\cite{zhang2017novel} & ECFP-14, molecular descriptors & \ding{51} & \ding{55} & Naive Bayes & 0.84 \\
        \textbf{Hansen et al.}\cite{hansen2009benchmark} & Molecular Descriptors & \ding{51} & \ding{55} & Support Vector Machine & 0.86 \\
        \textbf{Karim et al.}\cite{karim2019efficient} & 2D descriptors & \ding{51} &  \ding{55} & Neural Network, Decision Tree & 0.878\\
        \textbf{MutagenPred-GCNN}\cite{li2021mutagenpred} & Molecular graph & \ding{51} & \ding{51} & GCNN & 0.878\\\\
        \textbf{Shinada et al.}\cite{shinada2022optimizing} & ECFP4, molecular properties, genotoxicity alerts & \ding{51} & \ding{55} & Neural Network & 0.93\\
        \textbf{Our Method} & Fingerprints, 2D, 3D descriptors, and Molecular graph & \ding{51} & \ding{51} & SVM, RF, LGBM, CB, Extra Trees, GAT & \textbf{0.9521}\\\hline  
    \end{tabular}}
    \label{tab:sota}
\end{table}

\section{Ablation study}
We conducted an ablation study with the primary goal of systematically analyzing the contributions of individual modalities within our model. 
We assessed the impact of individual features on the overall performance of the model. 
This comprehensive analysis served to further evaluate the model capacity for generalization and its overall robustness. 
To investigate the impact, we investigated six feature combinations that included both single modality like STEM-FP, STEM-Desc, STEM-GAT and multiple modalities such as STEM-FPDesc, STEM-FPGAT, STEM-GATDesc. The details about each are given below and the results are shown in Table \ref{tab:ablation study} and \ref{tab:ablation study_toxcsm} for both datasets, respectively. 
\begin{itemize}
    \item STEM-FP: Employs the three fingerprints CDKExtended, KlekotaRoth and RDKit fingerprints as input for prediciting mutagenicity.
    \item STEM-Desc: Uses the 1D, 2D and 3D descriptors as input.
    \item STEM-GAT: Utilizes only the molecular graph embeddings as input for predicting mutagenicityy.
    \item STEM-FPGAT:  Uses the three fingerprints and molecular graph embeddings as input.
    \item STEM-FPDesc: Applies all three fingerprints and descriptors.
    \item STEM-GATDesc: Uses the molecular graph embeddings and descriptors as input for prediction.
\end{itemize}

From Table \ref{tab:ablation study}, it can be observed that the model STEM-FP, produces an accuracy and AUC of only 81.47\% and 78.13\% among all. The performances of STEM-Desc and STEM-GAT are not outstanding but close to each other showing that our GAT model is as strong as the widely used molecular descriptors.  However, an improvement in performance is evident as the features are integrated. For instance, STEM-FPDesc achieves an accuracy and AUC of 87.16\% and 95.0\%, respectively, representing a 3.71\%, 5.65\% and 5.71\%, 5.17\% increase over their individual results. 

\begin{table}[htbp]
    \centering
    \caption{The impact of various representations on the proposed model for the Hansen benchmark dataset}
    \scalebox{0.7}{
    \begin{tabular}{c c c c c c c c c c} \\\hline
         \textbf{Method} & \textbf{FP} & \textbf{Desc} & \textbf{GAT} & \textbf{Accuracy} & \textbf{Precision} & \textbf{Recall} & \textbf{F1-score} & \textbf{AUC} & \textbf{AUPR}  \\\hline\\
        STEM-FP & \ding{51} & \ding{55}  & \ding{55} & 0.8147 & 0.84 & 0.823 & 0.8278 & 0.8986 & 0.7813\\
        STEM-Desc &  \ding{55} & \ding{51}  & \ding{55} & 0.8345 & 0.8477 & 0.8462 & 0.847 & 0.8938 & 0.8006\\
        STEM-GAT &  \ding{55} & \ding{55}  & \ding{51}  & 0.8358 & 0.8353 & 0.868 & 0.8512 & 0.9034 & 0.7963\\
        STEM-FPDesc &  \ding{51}  & \ding{51}  & \ding{55} & 0.8716 & 0.8746 & 0.8903 & 0.8825 & 0.9503 & 0.8381\\
        STEM-FPGAT & \ding{51}  & \ding{55}  & \ding{51} & 0.8411 & 0.8438 & 0.867 & 0.8552 & 0.915 & 0.8035\\
        STEM-GATDesc & \ding{55} & \ding{51}   & \ding{51}  & 0.8848 & \textbf{0.8858} & 0.9018 & 0.8946 & 0.9436 & 0.8526\\\hline
        STEM  & \ding{51}  & \ding{51}   & \ding{51}  & \textbf{0.8868} & 0.8855 & \textbf{0.907} & \textbf{0.8966} & \textbf{0.9521} & \textbf{0.8544}\\
        \hline
    \end{tabular}}
    \label{tab:ablation study}
\end{table}
Likewise, STEM-GATDesc also outperforms STEM-Desc and STEM-GAT across all parameters. It causes an increment of 5.2\%, 5.03\%, 3.81\%, and 6.08\% compared to STEM-Desc in terms of AUC, accuracy, precision and recall, respectively. This finding suggests that the topological information acquired by molecular graph embeddings and the physical properties given by descriptors are complementary. Among all models, STEM-GATDesc has the highest precision of 0.8858. However, STEM, which integrates all three types of features, produces the most favorable results. Similar results are observed with Xu dataset as depicted in Table \ref{tab:ablation study_toxcsm}. This supports our notion that when substructural, topological, and physicochemical information are integrated, they have a synergistic impact, yielding optimal results.
\begin{table}[htbp]
    \centering
    \caption{The impact of various representations on the proposed model for Xu dataset\cite{xu2012silico}}
    \scalebox{0.7}{
    \begin{tabular}{c c c c c c c c c c} \\\hline
         \textbf{Method} & \textbf{FP} &  \textbf{Desc} & \textbf{GAT} & \textbf{Accuracy} & \textbf{Precision} & \textbf{Recall} & \textbf{F1-score} & \textbf{AUC} & \textbf{AUPR}  \\\hline\\
         STEM-FP & \ding{51}  & \ding{55}  & \ding{55} &  0.8754 & 0.8892 & 0.9059 & 0.8974 & 0.9391 & 0.8621\\
        STEM-Desc &  \ding{55} & \ding{51}   & \ding{55} & 0.8724 & 0.8848 & 0.9059 & 0.8952 & 0.9378 & 0.8582\\
         STEM-GAT &  \ding{55} & \ding{55} & \ding{51}  & 0.8383 & 0.8544 & 0.8817 & 0.8678 & 0.9143 & 0.8245 \\
         STEM-FPDesc &  \ding{51}   & \ding{51}  & \ding{55} & \textbf{0.878} & \textbf{0.8938} & 0.9018 & 0.8976 & 0.939 & 0.861\\
        STEM-DescGAT &  \ding{55} & \ding{51}   & \ding{51}  &  0.8687 & 0.8776 & 0.9086 & 0.8928 & 0.935 & 0.8524\\
         STEM-FPGAT & \ding{51}  & \ding{55} & \ding{51}  &  0.8694 & 0.8821 & 0.9038 & 0.8928 & 0.9384 & 0.8551\\
        
        \hline
         STEM  & \ding{51}  & \ding{51}   & \ding{51}  & 0.8765 & 0.8905 & \textbf{0.9062} & \textbf{0.8983} & \textbf{0.942} & \textbf{0.8635}\\
        \hline
    \end{tabular}}
    \label{tab:ablation study_toxcsm}
\end{table}

\section{Interpretation using SHAP}
\begin{figure}
     \centering
     \begin{subfigure}[b]{0.48\textwidth}
         \centering
         \includegraphics[width=\textwidth]{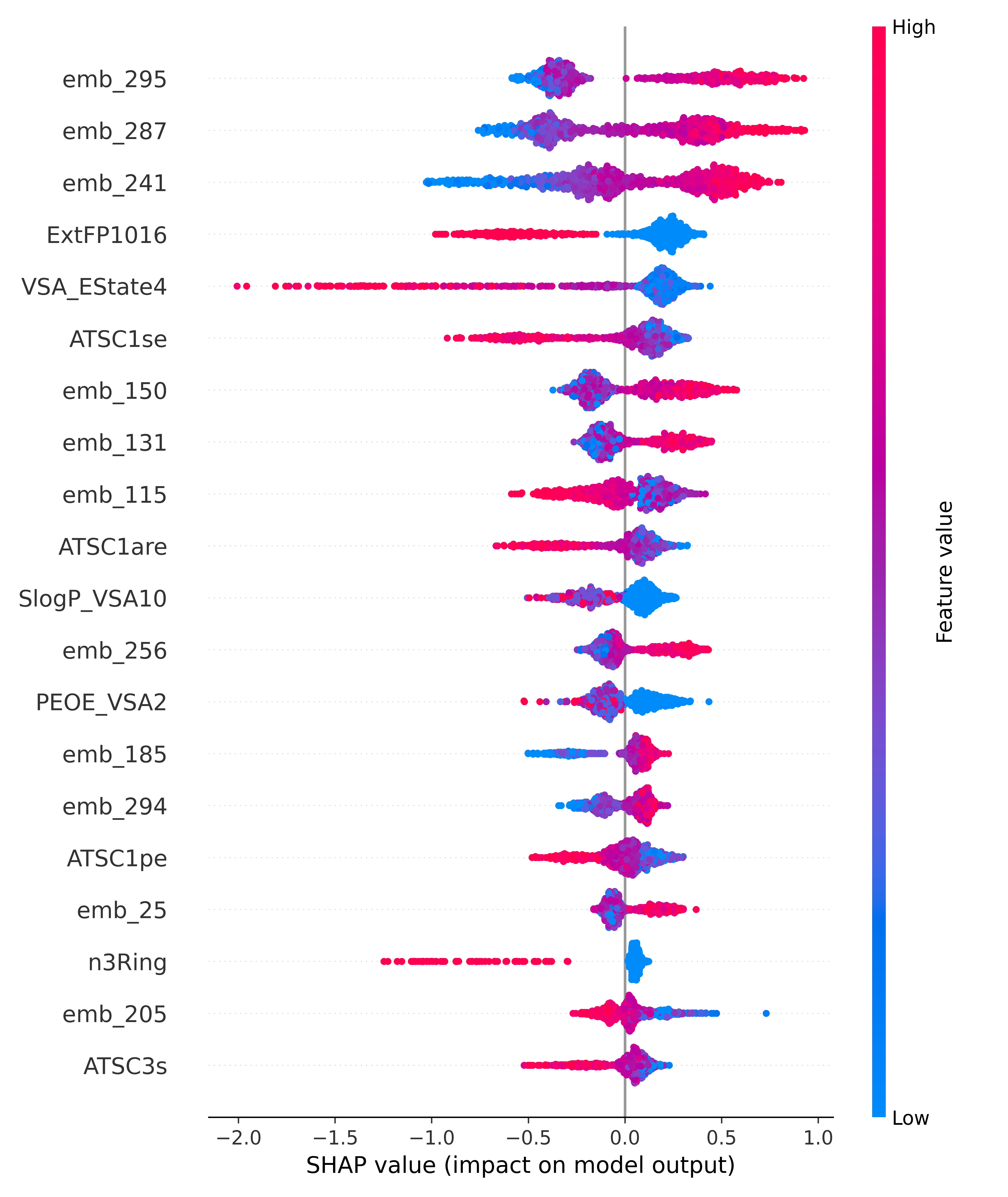}
          \caption{}
         \label{fig: shapsummary_lgbm}
     \end{subfigure}
     \begin{subfigure}[b]{0.48\textwidth}
         \centering
         \includegraphics[width=\textwidth]{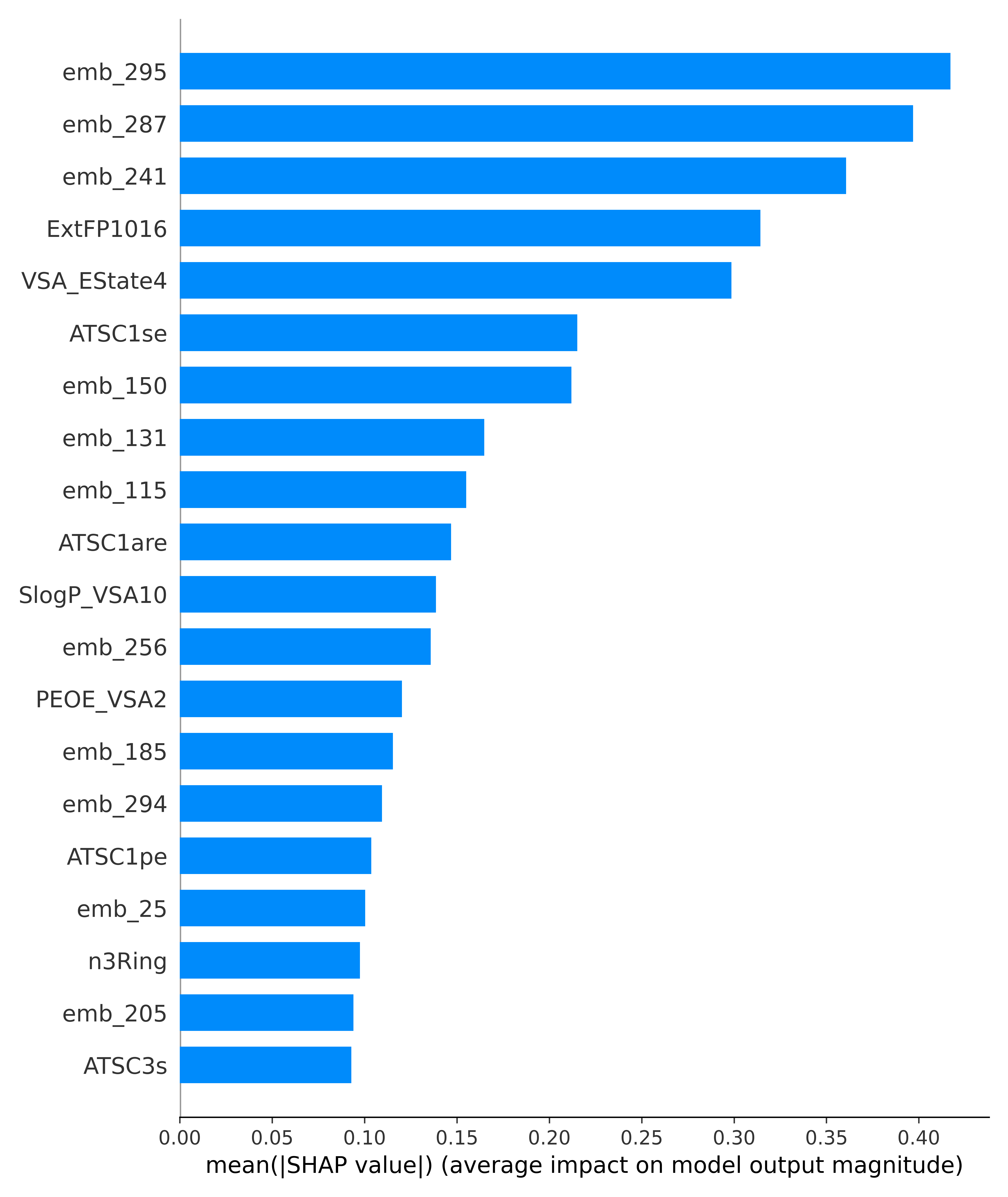}
         \caption{}
         \label{fig:shapbar_lgbm}
     \end{subfigure}
        \caption{The shap plots of the LGBM classifiers  (a)  Summary plot and  (b) Feature importances}
        \label{fig: SHAP_lgbm}
\end{figure}

The ultimate stacked model possesses a ‘black-box’ nature due to the integration of multiple classifiers. To address this, we have incorporated the SHAP method to elucidate the underlying rationale of the model predictions. The SHAP method serves the purpose of quantifying the influence of each classifier predictions on the stacked model determining the ranking in terms of their contribution to the final stacked model.
\begin{figure}[ht!]
     \centering
     \begin{subfigure}[b]{0.48\textwidth}
         \centering
         \includegraphics[width=\textwidth]{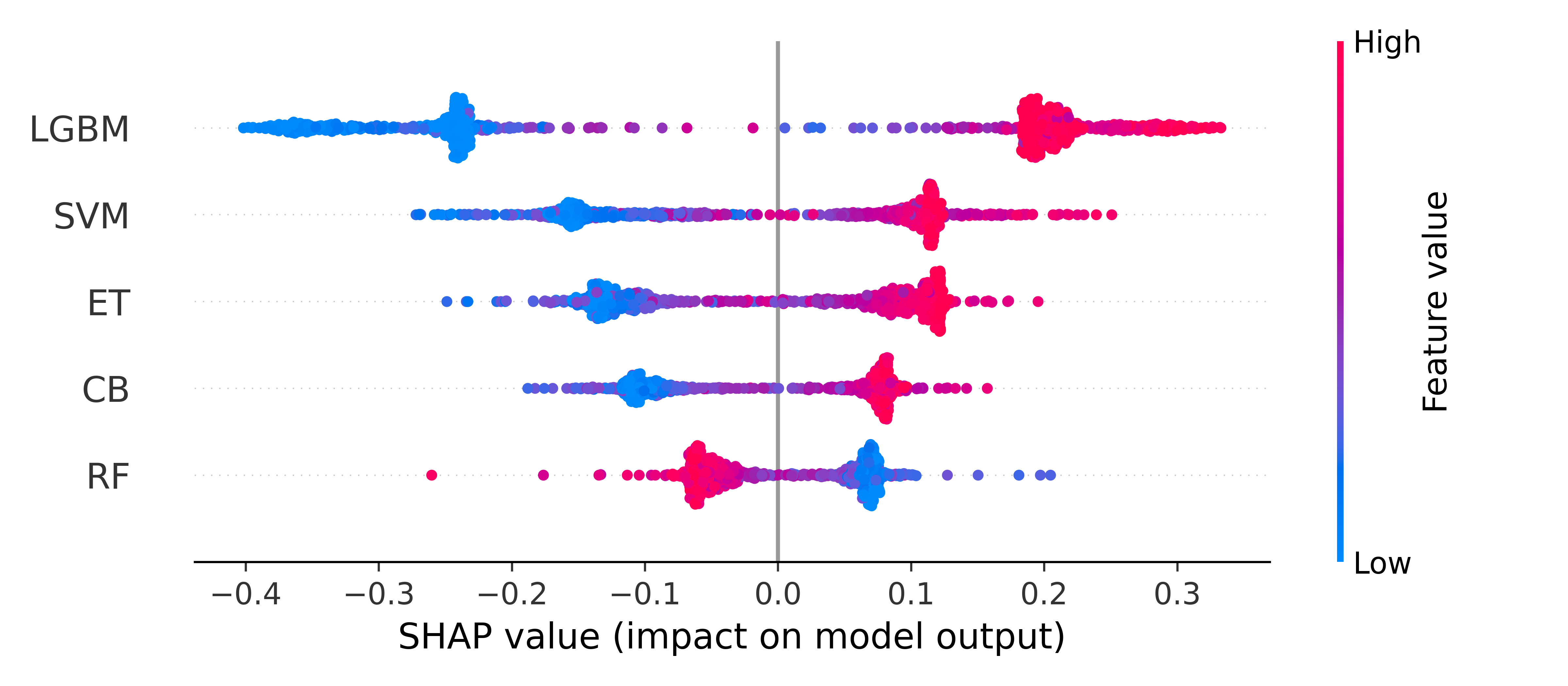}
          \caption{}
         \label{fig: shapsummary}
     \end{subfigure}
     \begin{subfigure}[b]{0.48\textwidth}
         \centering
         \includegraphics[width=\textwidth]{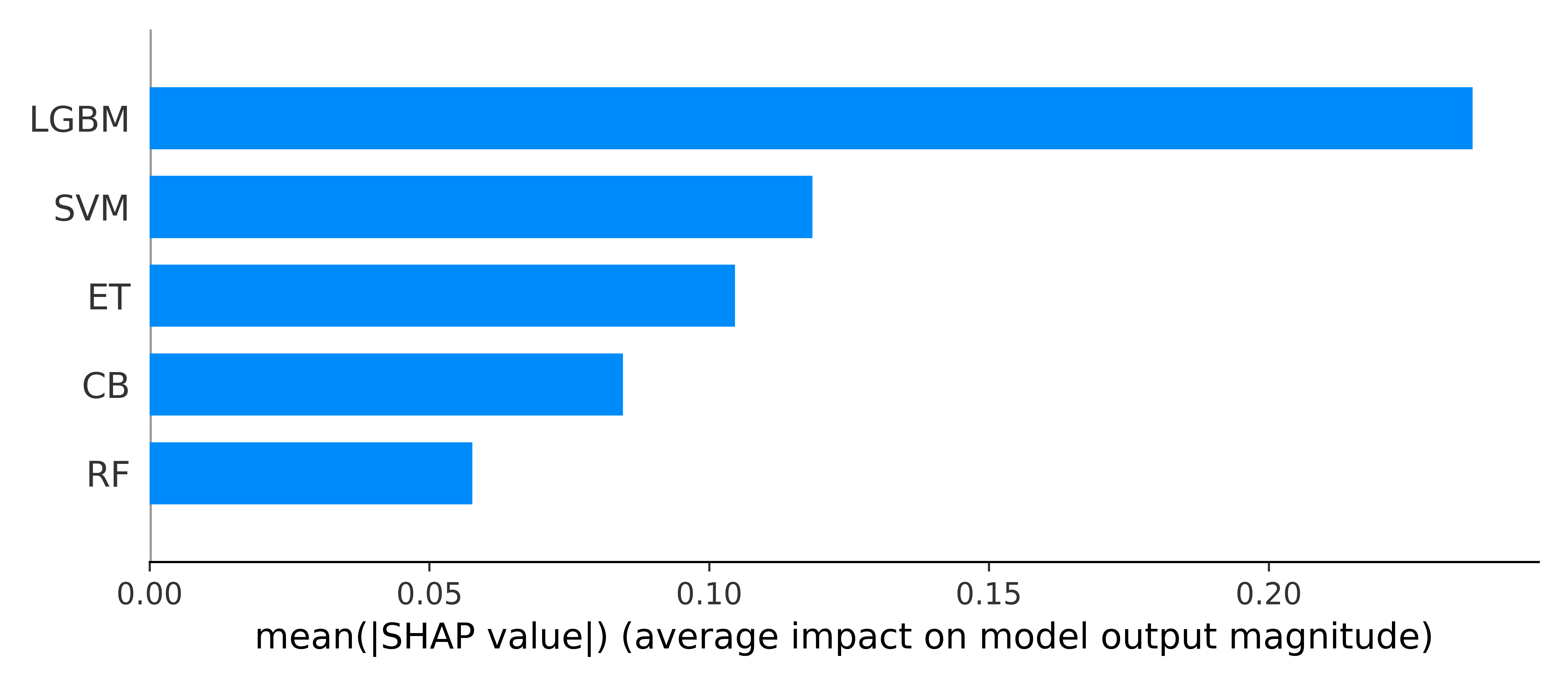}
         \caption{}
         \label{fig:shapbar}
     \end{subfigure}
        \caption{The shap plots for the classifiers  (a)  Summary plot and  (b) Importance of each classifier}
        \label{fig: SHAP}
\end{figure}
Figure \ref{fig: SHAP} depicts the summary and significance plots for the hansen data, where each row represents a classifier and each dot represents a sample. The predictions made by the individual classifiers served as the features. The SHAP values are represented on the X-axis, with positive and negative values indicating an increasing and decreasing effect on the predictions.
It can be clearly observed that the predictions made by the LGBM contributes the most to our stacked model. Subsequently, the SVM predictions rank as the second most influential, succeeded by ET, CB, and RF in descending order of importance. 
Given the paramount contribution of LGBM, we delve deeper into the validation by assessing feature importances through SHAP, as illustrated in Figure \ref{fig: SHAP_lgbm}. Figure shows that feature emb\_295, which is the embeddings of GAT at position 295 has positive impact on the prediction. 
The features with the higher values may probably give greater contributions to the  prediction. It can be observe that emb\_295 is the most important feature followed by emb\_287, emb\_241 and ExtFP1061 (CDKExtended fingerprint at 1061 position), etc.

\section{Analysis of structural alerts}

\begin{figure}[ht!]
     \centering
     \includegraphics[width=1.0\textwidth, height=6cm]{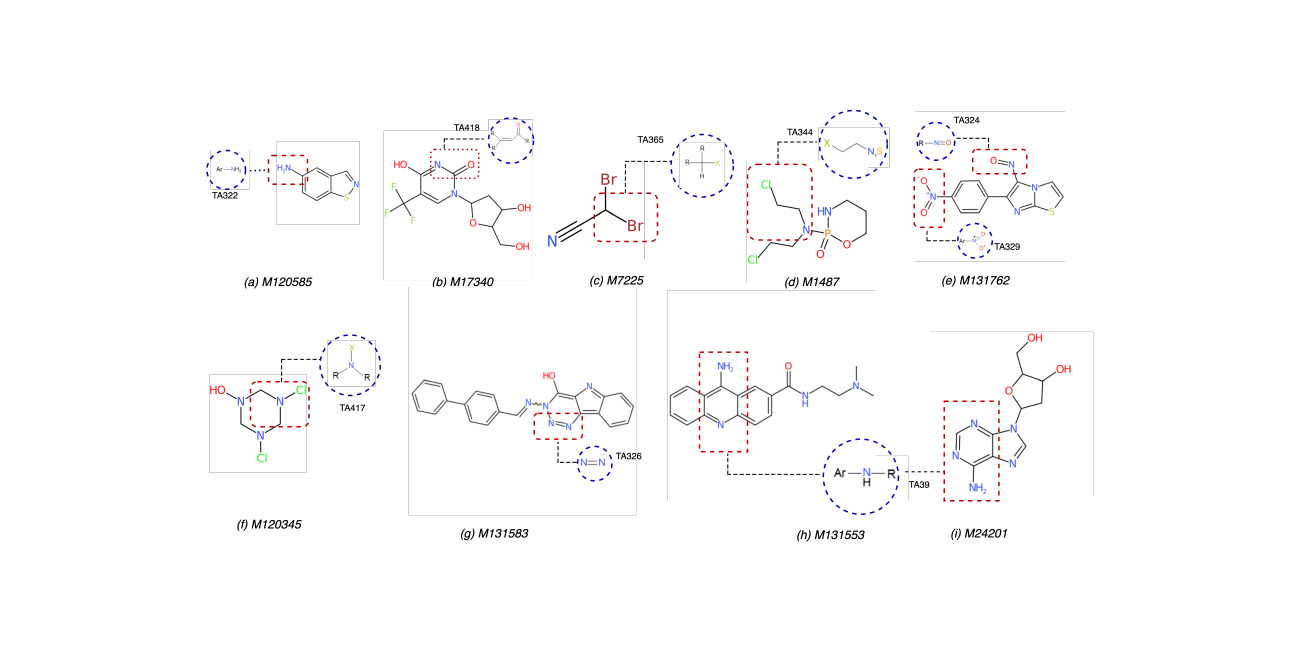}
        \caption{The analysis of structural alerts of  mutagens predicted by STEM}
        \label{fig: SA}
\end{figure}

In this section, we aimed to further demonstrate the effectiveness of STEM and illustrate its ability to predict mutagens while uncovering toxicophores. 
We independently predicted the compounds in the test set using STEM, without any reliance on the training set. Subsequently, we compared these predicted mutagens with the confirmed structural alerts provided by ToxAlerts \cite{sushko2012toxalerts}. The results illustrate that STEM successfully identified general toxicophores, including aromatic amines (TA322), Azotype compounds (TA326), N-chloroamines (TA436), and nitroso groups (TA324), as depicted in the figure \ref{fig: SA}. 
Among these structural alerts (SAs), TA322 is recognized as a symbolic toxicophore associated with mutagenicity, and the mutagenic mechanism of compounds containing TA322, TA436, TA326, and TA324 has been thoroughly elucidated and validated \cite{fishbein2011potential}. 
STEM demonstrates its capability not only to identify general structural alerts (SAs) but also to pinpoint specific SAs, such as aromatic nitro (TA329), nitrogen and sulfur mustard (TA344), and aliphatic halide (TA342). TA344 exhibits significant intrinsic reactivity and has been established as a specific toxicophore \cite{kazius2005derivation}. 
As depicted in Figure \ref{fig: SA}(f)-(i), certain compounds initially classified as non-mutagens were predicted by STEM to contain structural alerts (SAs) associated with mutagenicity. This observation highlights capability of STEM to classify compounds based on the presence of specific SAs, suggesting its potential to accurately assess the mutagenic potential of compounds based on structural features.

\section{Discussion and Conclusion}
In this study, we 
create a mutagenicity prediction model STEM based on multiple modalities. We use 5-fold CV strategy with 10 repeats on different seed values to thoroughly assess the model capabilities. Our model integrates heterogeneous molecular information and achieves classification via a stacked ensemble model achieving an AUC of 95.21\%. The remarkable predictive performance of STEM can be attributed to multiple factors. The use of GAT in particular facilitates the capture of complex spatial information within molecules. Furthermore, by merging substructural and physicochemical data in a synergistic manner, STEM creates a comprehensive picture of molecular structures, improving its predictive accuracy and robustness.
Nevertheless, STEM exhibits certain limitations that warrant further improvement. Firstly, considering a diverse array of ML classifiers beyond the current selection could enhance the construction of a more robust stacked model. Secondly, the exploration of additional fingerprints may contribute to a more nuanced representation of the molecular structure. Thirdly, investigating alternative variants of graph neural networks could provide a more comprehensive approach to capturing spatial information. Therefore, future research could delve into the impact of different GNN variants on spatial information capture within this study.







\end{document}